\documentclass{article}
\usepackage{spconf,amsmath,graphicx}
\usepackage{amssymb}
\usepackage{multirow}
\usepackage{graphics}
\usepackage{adjustbox}
\usepackage{booktabs}
\usepackage{tabularx}
\usepackage{enumitem}
\usepackage{xcolor}
\usepackage{caption}
\usepackage{subcaption}

\title{Towards the Detection of AI-Synthesized Human Face Images}

\name{Yuhang Lu and Touradj Ebrahimi\thanks{Support from the Swiss National Science
Foundation (SNSF) 20CH21\_195532 for XAIface
CHIST-ERA-19-XAI-011 is acknowledged.}}
\address{Multimedia Signal Processing Group (MMSPG)\\
\'Ecole Polytechnique F\'ed\'erale de Lausanne (EPFL)}

\begin{document}
%
\maketitle
\begin{abstract}
Over the past years, image generation and manipulation have achieved remarkable progress due to the rapid development of generative AI based on deep learning. Recent studies have devoted significant efforts to address the problem of face image manipulation caused by deepfake techniques. However, the problem of detecting purely synthesized face images has been explored to a lesser extent. In particular, the recent popular Diffusion Models (DMs) have shown remarkable success in image synthesis. Existing detectors struggle to generalize between synthesized images created by different generative models. In this work, a comprehensive benchmark including human face images produced by Generative Adversarial Networks (GANs) and a variety of DMs has been established to evaluate both the generalization ability and robustness of state-of-the-art detectors. Then, the forgery traces introduced by different generative models have been analyzed in the frequency domain to draw various insights. The paper further demonstrates that a detector trained with frequency representation can generalize well to other unseen generative models.

\end{abstract}
\section{Introduction}
\label{sec:intro}

In recent years, rapid advances have been made in image manipulation and synthesis techniques, such as generative adversarial networks (GANs) \cite{karras2017progressive, zhu2017unpaired, karras2019style, Karras2019stylegan2, esser2021taming} and variational autoencoders (VAE) \cite{kingma2013auto}. In practice, these deep learning-based techniques facilitate the creation of counterfeit images or video by manipulating the face of a person, which refers to the popular term ``Deepfake''. The generated human face images are often too realistic to be distinguished by human observers, raising social trust concerns due to their potential exploitation for malicious purposes. Consequently, considerable efforts have been dedicated to detecting face manipulations and promising progress has been demonstrated~\cite{rossler2019faceforensics++,shiohara2022detecting,lu2023assessment}.

\begin{figure}[t]
	\centering
	\begin{adjustbox}{width=\linewidth}
    \includegraphics[]{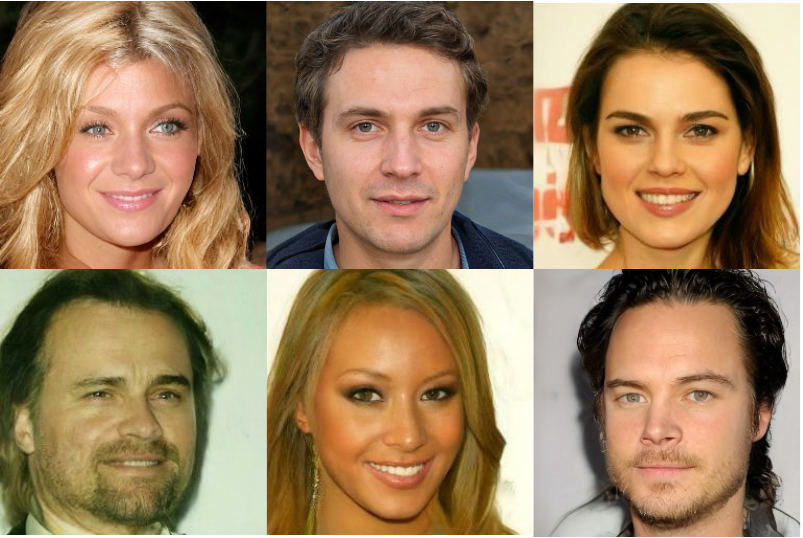}
	\end{adjustbox}
\caption{Realistic synthetic human face images generated by ProGAN \cite{karras2017progressive}, StyleGAN2 \cite{Karras2019stylegan2}, DDPM \cite{ho2020denoising}, DDIM \cite{song2020denoising}, PNDM \cite{liu2022pseudo}, and LDM \cite{rombach2022high} respectively.}
\label{fig:example}
\end{figure}

Nevertheless, another source of deepfake, i.e. entire face synthesis, has not received adequate attention so far. Various GAN-based models \cite{karras2017progressive, Karras2019stylegan2,esser2021taming} have been designed to create face images that do not exist in the world and produce surprisingly realistic results. More recently, the surge of Diffusion Models (DMs) has started a new paradigm in photorealistic image synthesis and an increasing number of researchers have been using them  to further improve the quality of produced results. With a publicly available model in the open-source community, one can easily create tons of fake human face images with little effort. 
Although this type of deepfake holds potential utility for applications such as video game character modeling, it can also be abused to create fake profiles for fraud or assist in spreading misinformation. 
Due to the diversity of different generative models, it remains a big challenge to develop a universal detection method that can identify synthetic face images created by arbitrary models.


Fortunately, a growing number of detection methods \cite{wang2019cnngenerated, marra2019gans, yu2019attributing, mandelli2020training, mandelli2022detecting, ojha2023towards, tan2023learning} have been developed for purely AI-synthesized images. Some rely on simple training of convolutional neural network (CNN) classifiers with various data preprocessing or augmentation strategies \cite{wang2019cnngenerated, mandelli2022detecting}, while others exploit specific fingerprints left by the generation techniques \cite{marra2019gans, yu2019attributing, tan2023learning}. 
Despite these advances, several concerns still persist in current studies. First, most of the detection methods only focus on images produced by a specific type of generative model. The generalization ability of such detectors to images created by different GAN models or recent diffusion models is not sufficiently studied. Although recent studies \cite{ricker2022towards, corvi2023detection} have made preliminary progress in the right direction, their focus has predominantly centered on general categories of synthetic images with rich contextual information, such as bedrooms, outdoor churches, etc. This brings a second concern, i.e. whether detectors for generic fake images can perform well for synthetic human face images. Third, the resistance of a detector against common image perturbations, particularly on DM-generated face images, remains unexplored.




This paper addresses the challenges in detecting entirely AI-synthesized human face images. The primary contribution lies in the establishment of a new benchmark for this task, achieved by systematically generating a substantial volume of synthetic human face images using seven popular generative models.
Subsequently, the generalization ability and robustness of various learning-based detectors have been evaluated with the benchmark. 
The paper also aims to draw new insights for developing more generalizable detectors. To that end, a frequency domain analysis on the synthetic face images is carried out, examining the deviation of their spectra from that of real images. Consequently, our experimental results demonstrate that training a learning-based detector using frequency representations yields outstanding performance and generalization ability in the benchmark.




\section{Related Work}



\subsection{Generative Models for Image Synthesis}
Generative adversarial networks (GANs) have long stood as the prevailing approach for numerous image synthesis tasks. In general, a GAN \cite{goodfellow2014generative} is trained through a competing game between two models, i.e., a generator and a discriminator. The generator aims to fool the discriminator by producing images resembling those in the training data, while the latter seeks to distinguish between real and generated images. In practice, some GAN models~\cite{karras2017progressive, karras2019style} take noise as input and are able to generate high-resolution images with good perceptual quality, while others~\cite{zhu2017unpaired, isola2017image} are conditioned on additional information, such as a semantic map or another image, often employed for translation between two images.  
This paper focuses on unconditional face image generation and adopts three GAN models that are pre-trained on high-quality face image datasets, namely ProGAN \cite{karras2017progressive}, StyleGAN2 \cite{Karras2019stylegan2}, and VQGAN \cite{esser2021taming}.

More recently, initially inspired by non-equilibrium thermodynamics \cite{sohl2015deep}, diffusion models have become a new paradigm for image generation. Ho et al. \cite{ho2020denoising} proposed denoising diffusion probability models (DDPM) and showed an impressive ability in image synthesis in comparison to GAN-based counterparts.
Song et al. \cite{song2020denoising} explored the use of the denoising diffusion implicit model (DDIM) to improve sampling speed while maintaining good image quality. ADM~\cite{dhariwal2021diffusion} introduced a more effective architecture incorporating classifier guidance and demonstrated superior performance when compared to GANs. Liu et al. \cite{liu2022pseudo} proposed PNDM and further enhanced the sampling efficiency and generation quality. A later work LDM \cite{rombach2022high} integrated text and image inputs in latent space via a cross-attention mechanism. 
When these diffusion models are trained on large-scale human face datasets, they are capable of generating realistic and high-quality face images. This paper includes four diffusion models in the benchmark, namely DDPM, DDIM, PNDM, and LDM.


\subsection{Detection of AI-Synthesized Images}
The need for fake image detectors has existed ever since the appearance of various generative models.  
Some detection methods leveraged hand-crafted features, such as color cues~\cite{mccloskey2018detecting}, saturation cues \cite{mccloskey2019detecting}, blending artifacts \cite{li2020face}, and gradients \cite{tan2023learning}, while other studies relied on CNN-based classifiers to detect fake images. 
Several researchers have  leveraged various advanced neural network architectures as primary solutions. For example, R\"ossler et al. \cite{rossler2019faceforensics++} retrained XceptionNet~\cite{chollet2017xception} with a large-scale deepfake dataset. Cozzolino et al.~\cite{cozzolino2018forensictransfer} learned a forensic embedding through an autoencoder-based architecture to distinguish between real and fake images and performed well on StyleGAN-generated images. Marra et al. \cite{marra2018detection} tested multiple CNN-based architectures for detecting GAN-generated images.

However, most of the detection methods above only show good performance when the fake images share the same distribution as the training data. This is why, more attention has been recently devoted to the detector's generalization ability. 
Wang et al.~\cite{wang2019cnngenerated} proposed to train a basic detection network with data preprocessed by JPEG compression and Gaussian Blur, and surprisingly generalized well on other unseen GAN-generated images. Grag et al. \cite{gragnaniello2021gan} improved based on \cite{wang2019cnngenerated} by updating the network architecture. 
Shiohara et al. \cite{shiohara2022detecting} fine-tuned a pre-trained EfficientNetB4 \cite{tan2019efficientnet} to detect blending boundary artifacts and achieved promising results in cross-data evaluation on several deepfake benchmarks. 
Mandelli et al. \cite{mandelli2022detecting} leveraged an ensemble of multiple EfficientNetB4 that were trained under different conditions and achieved the state-of-the-art. 

An increasing number of studies have been carried out to counteract the emerging realistic fake images created by diffusion models. DIRE \cite{wang2023dire} developed an effective method to detect DM-generated images by reconstructing an input image through a pre-trained diffusion model. 
Lorenz et al. \cite{lorenz2023detecting} exhibited the superiority of multi-local intrinsic dimensionality in diffusion detection. Ojha \cite{ojha2023towards} proposed a universal fake image detector by leveraging a pre-trained large vision-language model and achieved excellent generalization ability across GAN and DM-based fake images. 


 

\begin{table}[t]
  \centering
  \caption{Generative models used in this work, including three GAN models and four diffusion models. The quality of a face dataset produced by each model is reported with FID scores. 10k images are randomly sampled for each model to calculate FID score. A lower FID refers to higher quality.}
  \resizebox{\linewidth}{!}{
    \begin{tabular}{cccc}
    \toprule
    Model Family & Method & Publication & FID\\
    \midrule
    \multirow{3}[2]{*}{GANs} & ProGAN & Karras et al. (2018) \cite{karras2017progressive} & 12.39 \\
          & StyleGAN2 & Karras et al. (2019) \cite{Karras2019stylegan2} & 15.17 \\
          & VQGAN & Esser et al. (2021) \cite{esser2021taming} & 12.99 \\
    \midrule
    \multirow{4}[2]{*}{DMs} & DDPM  & Ho et al. (2020) \cite{ho2020denoising} & 16.64\\
          & DDIM  & Song et al. (2020) \cite{song2020denoising} & 14.36\\
          & PNDM  & Liu et al. (2022) \cite{liu2022pseudo} & 13.97\\
          & LDM   & Rombach et al.(2022) \cite{rombach2022high} & 7.28\\
    \bottomrule
    \end{tabular}%
    }
  \label{tab:data}%
\end{table}%

\section{Detection Benchmark for Synthetic Human Face Images}
This paper contributes a comprehensive benchmark for synthetic face image detection. This section first introduces the collected dataset along with the detectors incorporated into the benchmark. Then, two major objectives of the benchmark, i.e., evaluating the generalizability and robustness of a detector, and how to achieve them are elaborated.




\begin{figure*}[t]
\centerline{\includegraphics[width=0.8\linewidth]{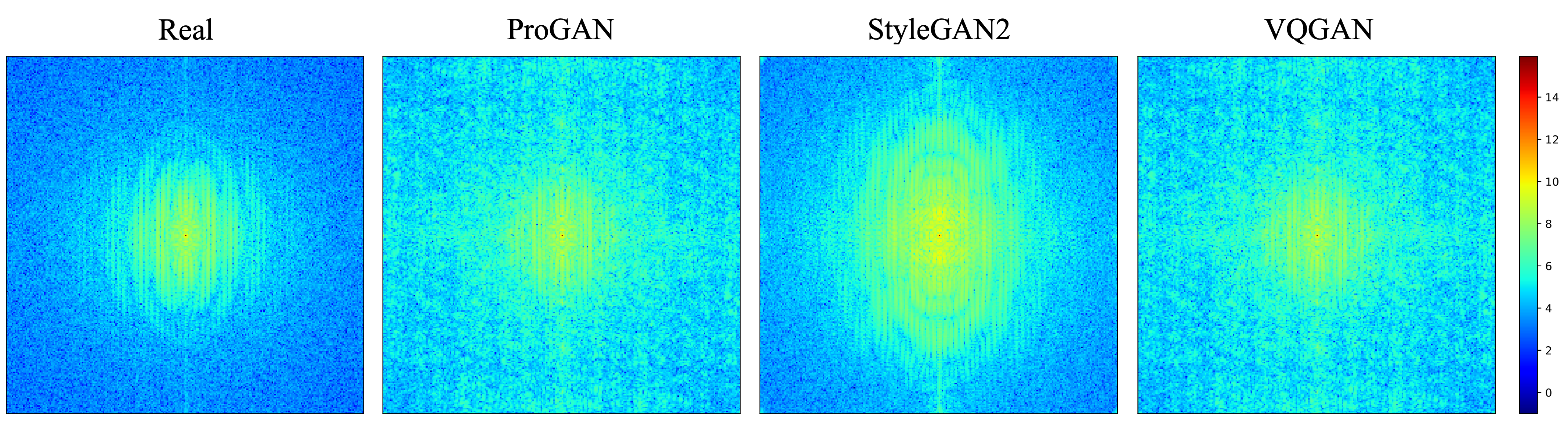}}
\caption{Mean frequency spectra of real images from CelebA-HQ \cite{karras2017progressive} and synthetic human face images created by three GAN models, namely ProGAN \cite{karras2017progressive}, StyleGAN2 \cite{Karras2019stylegan2}, and VQGAN \cite{esser2021taming}.}
\label{fig:freq1}
\end{figure*}

\begin{figure*}[t]
\centerline{\includegraphics[width=\linewidth]{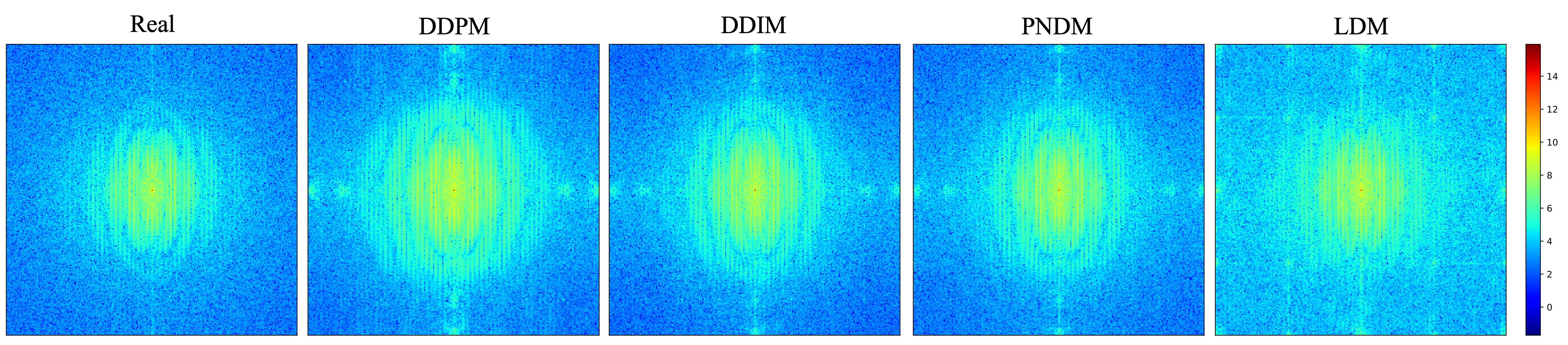}}
\caption{Mean frequency spectra of real images from CelebA-HQ \cite{karras2017progressive} and synthetic human face images created by four diffusion models, namely DDPM \cite{ho2020denoising}, DDIM \cite{song2020denoising}, PNDM \cite{liu2022pseudo}, and LDM \cite{rombach2022high}.}
\label{fig:freq2}
\end{figure*}

\subsection{A Dataset of Synthetic Face Images}

This work first collects a dataset that comprises real images from the CelebA-HQ \cite{karras2017progressive} dataset and synthetic human face images created by seven cutting-edge generative models. As shown in Table \ref{tab:data}, the fake face images are synthesized by GANs, including ProGAN \cite{karras2017progressive}, StyleGAN2 \cite{Karras2019stylegan2}, and VQGAN \cite{esser2021taming}, and DMs, including DDPM \cite{ho2020denoising}, DDIM \cite{song2020denoising}, PNDM \cite{liu2022pseudo}, and LDM \cite{rombach2022high}. For LDM, the unconditional mode is employed because bimodal inputs (e.g. incorporating texts) will result in unnatural face images. More specifically, StyleGAN2 is pre-trained on the FFHQ \cite{karras2019style} dataset and all other models are pre-trained on CelebA-HQ \cite{karras2017progressive} dataset. The default resolution for the entire dataset is set to 256$\times$256 because it is the most common output size among the selected generative models. Higher-resolution images generated by certain models are downscaled to 256$\times$256 using bilinear interpolation. Under these settings, all models are capable of generating realistic human face images, see examples in Figure \ref{fig:example}. The Fr\'echet inception distances (FID) reported in Table \ref{tab:data} further show that all models produce images of comparable quality. 

For each generation technique, 40k images are collected in total and they are by default split into 38k, 1k, and 1k for training, validation, and testing purposes. 


\subsection{Detectors}
Several learning-based methods for synthetic image detection are selected for experiments in the benchmark. All methods report satisfactory performance in prior studies on general fake image detection tasks. However, their performance specifically concerning synthetic human face images, their adaptability to DM-generated images, and their robustness against common perturbations have not been investigated until this paper. The selected detectors are outlined as follows.

Wang2020 \cite{wang2019cnngenerated} employed a ResNet-50 architecture and trained with JPEG compression and Gaussian blurring as augmentation, obtaining fair generalization ability among GAN-synthesized images. Grag2021 \cite{gragnaniello2021gan} built upon this approach by further exploring different variations of ResNet-50 to enhance performance in real-world scenarios. 
Mandelli2022 \cite{mandelli2022detecting} used an ensemble of five orthogonal EfficientNetB4 \cite{tan2019efficientnet} networks to detect fake images. Each model was trained on different datasets created by various GAN models and augmented using different techniques. This strategy significantly improved the overall performance and generalization ability. Ojha2023 \cite{ojha2023towards} leveraged a large pre-trained vision-language model and exhibited exceptional generalization ability in detecting fake images across a variety of generative models. 


\subsection{Generalizability}
One of the main objectives of the proposed benchmark is to assess the generalization ability of a detector in the presence of synthetic human face images. This can be interpreted in one of the following two ways: (i) whether a detector trained with other categories of fake images can effectively generalize to synthetic human face images; (ii) whether a detector trained on images created by a specific generative model can still obtain satisfactory performance on unseen GANs and DMs. 
To address the first way, this work employs open-source detection methods pre-trained on various categories of synthetic images, such as bridge, church, etc., sourced from the LSUN dataset \cite{yu2015lsun}. They are directly evaluated by the synthetic face images from the benchmark. 
For the latter, training data is selected from one GAN model and one diffusion model in the benchmark and used to train detectors from scratch. Then, they are tested using the out-of-distribution fake face images synthesized by other GANs and DMs. 


\subsection{Robustness against Image Perturbation}
Synthetic face images often undergo various processing operations before dissemination, such as compression, resizing, etc. Therefore, in addition to the generalization ability, the proposed benchmark further measures the robustness of detectors against common image perturbations. Specifically, the impact of the following perturbations is analyzed:
\vspace{-5pt}
\begin{itemize}[leftmargin=*]
 \setlength\itemsep{-4pt}
    \item \textbf{JPEG compression} is performed and the impact of different quality factors are measured individually, i.e., \{10, 20, $\dots$, 90\}.
    \item \textbf{Blurry effect} is applied via a Gaussian Blur kernel. The kernel size is selected from \{3, 5, $\dots$, 15\}.
    \item \textbf{Gaussian noise} with zero mean is added and the standard deviation is selected from \{5, 10, $\dots$, 30\}. 
    \item \textbf{Resizing operation} is employed by first downsampling the image to lower resolutions by a scale of \{2, 4, $\dots$, 12\}, with bicubic interpolation and then upscaling to 256$\times$256.
\end{itemize}
Notably, only one type of perturbation of a fixed intensity is applied to the entire test set in each evaluation to avoid randomness. 

\begin{table*}[t]
  \centering
  \caption{Detection performance of four pre-trained detectors. The weights released by the original authors are utilized.}
    \resizebox{0.95\linewidth}{!}{
    \begin{tabular}{c|ccc|cccc|c}
    \toprule
    \multirow{2}[2]{*}{AUC/AP (\%)} & \multicolumn{3}{c|}{GANs} & \multicolumn{4}{c|}{DMs}      & \multirow{2}[2]{*}{Average} \\
    \cmidrule{2-8}
          & ProGAN & StyleGAN2 & VQGAN & DDIM  & DDPM  & PNDM  & LDM   &  \\
    \midrule
    Wang2020 \cite{wang2019cnngenerated} & 78.31/78.09 & 88.39/88.34 & 79.80/79.84 & 74.51/70.94 & 65.09/60.58 & 76.38/73.40 & 77.10/76.51 & 77.08/75.39\\
    Grag2021 \cite{gragnaniello2021gan} & 99.96/99.96 & 99.27/99.29 & 99.32/99.35 & 68.85/62.45 & 59.22/52.93 & 66.82/61.84 & 99.67/99.69 & 84.73/82.22 \\
    Mandelli2022 \cite{mandelli2022detecting} & 100.00/100.00 & 100.00/100.00 & 99.43/99.36 & 99.99/99.99 & 97.87/97.70 & 98.16/98.02 & 98.76/99.00 & 99.17/99.15 \\
    Ojha2023 \cite{ojha2023towards} & 96.38/96.52 & 73.24/69.57 & 96.20/96.27 & 97.78/97.97 & 93.31/93.57 & 96.37/96.55 & 98.18/98.19 & 93.42/92.66 \\
    \bottomrule
    \end{tabular}%
    }
  \label{tab:phase1}%
\end{table*}%

\begin{table*}[t]
  \centering
  \caption{Generalization analysis of various detection techniques. All methods are trained on face images generated by ProGAN and DDIM, and tested on images created by all seven generative models. To distinguish from prior experiments, $*$ here refers to the retrained version of Wang2020 and Ojha2023 using our training set. The best result is denoted by underscore.}
    \resizebox{\linewidth}{!}{
    \begin{tabular}{c|ccc|cccc|c}
    \toprule
    \multirow{2}[4]{*}{AUC/AP (\%)} & \multicolumn{3}{c|}{GANs} & \multicolumn{4}{c|}{DMs}      & \multicolumn{1}{c}{\multirow{2}[4]{*}{Average}} \\
\cmidrule{2-8}          & \multicolumn{1}{c}{ProGAN} & \multicolumn{1}{c}{StyleGAN2} & \multicolumn{1}{c|}{VQGAN} & \multicolumn{1}{c}{DDIM} & \multicolumn{1}{c}{DDPM} & \multicolumn{1}{c}{PNDM} & \multicolumn{1}{c|}{LDM} &  \\
    \midrule
    ResNet-50 & \underline{100.00}/\underline{100.00} & 52.95/50.52 & \underline{100.00}/\underline{100.00} & 61.60/61.07 & 73.32/71.25 & 60.02/60.93 & 37.97/43.86 & 69.41/69.66 \\
    XceptionNet & \underline{100.00}/\underline{100.00} & 57.65/57.21 & \underline{100.00}/\underline{100.00} & 53.59/59.16 & 52.26/51.30 & 57.51/59.16 & 39.94/44.65 & 65.85/67.35 \\
    EfficientNetB4 & \underline{100.00}/\underline{100.00} & 52.26/50.70 & \underline{100.00}/99.99 & 77.69/78.22 & 71.20/70.91 & 81.46/81.83 & 61.19/62.42 & 77.69/77.72 \\
    Wang2020$^*$ \cite{wang2019cnngenerated} & \underline{100.00}/\underline{100.00} & 52.41/52.98 & \underline{100.00}/\underline{100.00} & 90.28/90.34 & 85.17/84.42 & 86.97/87.27 & 64.77/65.52 & 82.80/82.93 \\
    Ojha2023$^*$ \cite{ojha2023towards} & 99.97/99.97 & 94.80/94.58 & 99.97/99.97 & \underline{99.72}/\underline{99.74} & 98.60/98.68 & 99.56/99.58 & \underline{99.94}/\underline{99.94} & 98.94/98.92 \\
    \midrule
    ResNet-50+FreqSpec & 99.79/99.78 & 93.60/92.96 & 99.84/99.83 & 98.38/98.37 & 98.77/98.59 & 99.30/99.43 & 99.55/99.49 & 98.46/98.35 \\
    XceptionNet+FreqSpec & 99.45/99.53 & 95.26/95.03 & 99.64/99.66 & 99.24/99.16 & 99.03/98.89 & 99.66/99.71 & 99.64/99.66 & 98.85/98.81 \\
    EfficientNetB4+FreqSpec & 99.87/99.89 & \underline{98.72}/\underline{98.68} & 99.95/99.95 & 99.53/99.53 & \underline{99.54}/\underline{99.53} & \underline{99.97}/\underline{99.97} & \underline{99.94}/\underline{99.94} & \underline{99.65}/\underline{99.64} \\ 
    \bottomrule
    \end{tabular}%
    }
  \label{tab:phase2}%
  \vspace{-3pt}
\end{table*}%


\section{Frequency Artifacts Analysis}
As the generation tools become more advanced, their results become indistinguishable from real images when observed by human subjects in the spatial domain and even some CNN-based detectors. Studies \cite{frank2020leveraging, wang2019cnngenerated, ricker2022towards, corvi2023detection} have identified characteristic fingerprints present in GAN-generated images via frequency analysis and observed grid-like artifacts in general categories of synthetic images.

This section analyzes forgery traces in the frequency domain, particularly for synthetic human face images created by various generative models. As suggested by prior work~\cite{wang2019cnngenerated}, each image is first converted to gray-scale by averaging over color channels and then high-pass filtered by subtracting a median-filtered version of itself. Subsequently, the Fast Fourier Transform (FFT) is applied to the processed image to extract the frequency spectrum, with magnitude values log-scaled for better visualization. Figure \ref{fig:freq1} and \ref{fig:freq2} depict the average frequency spectrum of 1,000 images randomly sampled from the real CelebA-HQ dataset and seven fake face image datasets created generative models listed in Table \ref{tab:data}.

As shown in Figure \ref{fig:freq1}, the common grid-like artifacts found in previous studies \cite{wang2019cnngenerated, frank2020leveraging} are notably absent in our GAN-generated face datasets. However, datasets created by ProGAN and VQGAN exhibit numerous high-frequency noises. The more advanced StyleGAN2 contains relatively fewer such artifacts but remains distinguishable from real image spectra. 
On the other hand, Figure \ref{fig:freq2} shows that the FFT spectra of DM-created face images closely resemble the real spectrum, except for LDM which contains both high-frequency noise and grid-form artifacts. While images produced by DDPM, DDIM, and PNDM exhibit fewer visible artifacts in the frequency domain, they tend to have higher spectra density and contain low-frequency artifacts along the vertical and horizontal impulse sequence, deviating from that of real image spectra. 

Observing notable discrepancies between real and synthetic face images in their frequency representations, this paper further explores the potential utility of these differences in training a more generic detector that can identify fake face images generated by various GANs and DMs. 
Specifically, the detection task is framed as a binary classification process and three basic classification networks are selected, namely ResNet-50 \cite{he2016deep}, XceptionNet \cite{chollet2017xception}, and EfficientNetB4 \cite{tan2019efficientnet}. These networks are trained with only frequency representations of both real and synthetic face images. Further details about experimental setups and results are presented in the next section. 




\section{Detection Performance}

\subsection{Experimental Setup}
Experiments are structured into three phases. In phase one, four popular detection methods are first evaluated on our benchmark, namely Wang2020 \cite{wang2019cnngenerated}, Grag2021 \cite{gragnaniello2021gan}, Ojha2023 \cite{ojha2023towards}, Mandelli2022 \cite{mandelli2022detecting}. The pre-trained weights directly released by the authors are used. The former three detectors were originally trained on general categories of synthetic images sourced from the LSUN dataset \cite{yu2015lsun}, while the latter was trained on a synthetic face dataset \cite{karras2021alias} for comparative analysis. The four detectors are tested with all seven test sets provided by the benchmark.


In the second phase, Wang2020 and Ojha2023 are selected to evaluate their generalization ability across various generative models. Additionally, three CNN classifiers (ResNet-50, XceptionNet, EfficientNetB4) and their counterparts trained with frequency representations are also evaluated under the same setting. In detail, all the detectors are trained on real images sourced from CelebA-HQ, and fake images created by one GAN model (ProGAN) and one diffusion model (DDIM). The number of real images is upsampled accordingly to ensure a balanced training set. Then, tests are performed under the same configuration as in phase one.


In phase three, the robustness of detectors against four common image perturbations is assessed. The evaluation involves~four pre-trained detectors from phase one, tested on distorted face images generated by ProGAN and DDIM. Furthermore, Wang2020 and Ojha2023 retrained on the corresponding training set are also incorporated and tested under the same conditions.

\subsection{Evaluation Metrics}
Following previous work about synthetic image detection and benchmarking, the average precision (AP) and Area Under Receiver Operating Characteristic Curve (AUC) scores are used to evaluate the detectors. 

\begin{figure*}[t]
     \centering
     \begin{subfigure}[b]{0.95\textwidth}
         \centering
         \includegraphics[width=\textwidth]{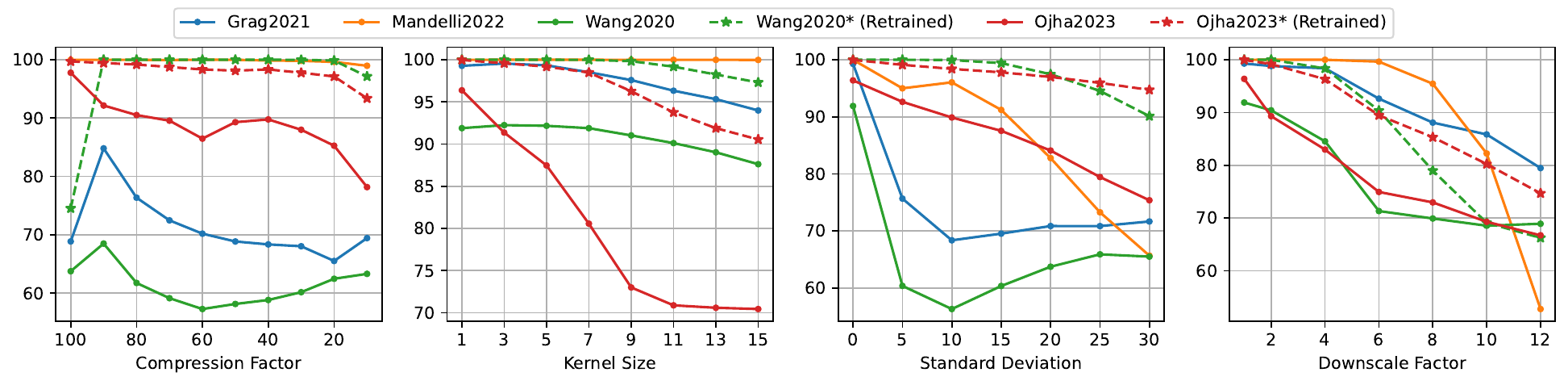}
         \caption{Evaluation performance on test set created by ProGAN.}
         \label{fig:robust1}
     \end{subfigure}
     \hfill
     \begin{subfigure}[b]{0.95\textwidth}
         \centering
         \includegraphics[width=\textwidth]{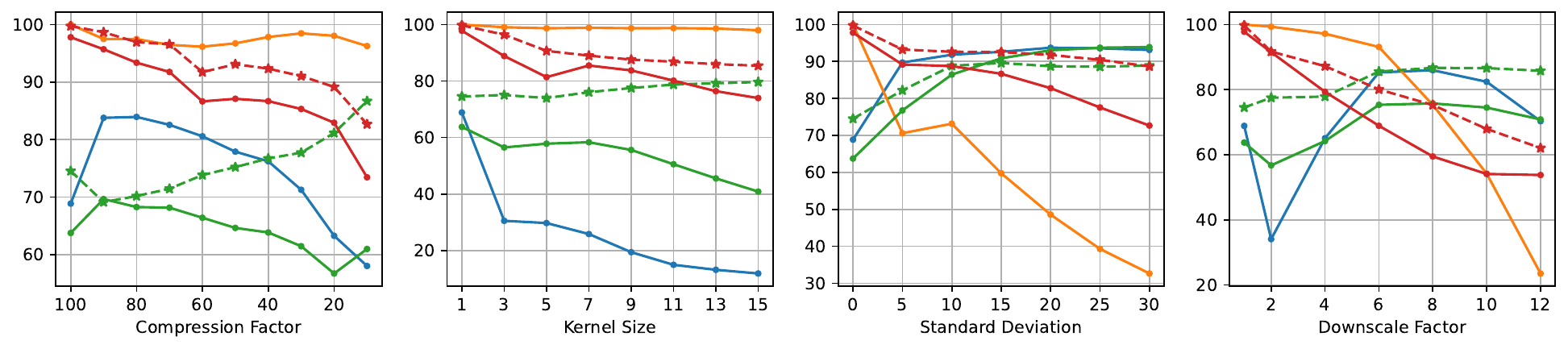}
         \caption{Evaluation performance on test set created by DDIM.}
         \label{fig:robust2}
     \end{subfigure}
     \hfill
     \caption{Performance of various detectors under the perturbation of JPEG compression, Gaussian blur, Gaussian noise, and resizing operation (from left to right). The evaluation is conducted on two test sets created by ProGAN and DDIM respectively.}
     \label{fig:robust}
\end{figure*}

\subsection{Experimental Results}
This section presents the results of the three-phase evaluation. 
First of all, the performance of four pre-trained detectors is reported in Table \ref{tab:phase1}. Wang2020 shows poor performance on both GAN and DM-generated face images, although in previous study it reported fair adaptability among general categories of fake images. Grag2021 is able to generalize among face images created by different GAN models but fails to achieve good performance on three diffusion models, i.e., DDPM, DDIM, and PNDM. Ojha2023 demonstrates good transferability across face images synthesized by most GANs and DMs, except for StyleGAN2. In comparison, the Mandelli2022 detector that has been trained on a GAN-created face dataset shows exceptional performance in our benchmark. To sum up, detectors only trained on general fake images struggle to adapt to synthetic face images.


Secondly, Table \ref{tab:phase2} summarizes detectors' performance after being trained with fake face images from the benchmark. 
As a result, the retrained version of Wang2020 shows the potential to generalize to images created by VQGAN, DDPM, and PNDM, yet struggles to adapt to StyleGAN2 and LDM. Conversely, Ojha2023 achieves nearly flawless detection across all the GANs and DMs. Notably, after training with frequency representations of these face images, the three CNN detectors achieve much better generalization ability when compared to their counterparts that are directly trained with RGB images. The combination of EfficientNetB4 and frequency representation even surpasses the state-of-the-art performance on certain GAN models and most DMs. 

Thirdly, Figure \ref{fig:robust} illustrates the robustness of various detectors under four common image perturbations. The four solid lines represent the performance of the four pre-trained detectors from phase one. 
Although both Wang2020 and Grag2021 incorporate data augmentation techniques, they are notably affected by compression artifacts and noise. Similarly, the performance of Ojha2023 deteriorates as the perturbation intensity increases. Mandelli2022 remains the most robust among the four, particularly in handling data subjected to JPEG compression and Gaussian blur effect. However, its performance inevitably declines in the presence of heavy noise or low-resolution effects. 
The two dashed lines additionally depict the performance of the retrained version of Wang2020 and Ojha2023 using the training set generated by ProGAN and DDIM. While the overall evaluation results improve, Figure \ref{fig:robust2} reveals that they are not resilient enough to compression artifacts and low-resolution effects. 

\section{Conclusion}

This paper addressed detection of entirely AI-synthesized human face images. A comprehensive benchmark was devised to assess fake image detectors in terms of adaptability and robustness. Results show that detectors only trained on general categories of fake images have difficulty generalizing to synthetic face images. The generalization across various GANs and DMs and robustness against perturbations also remain two important challenges in most detection methods. Furthermore, the paper examined forgery traces of synthetic face images in the frequency domain and demonstrated that training a detector with frequency representation can significantly enhance its performance and generalization ability.  




\let\oldbibliography\thebibliography
\renewcommand{\thebibliography}[1]{%
  \oldbibliography{#1}%
  \setlength{\itemsep}{-1pt}%
}

\bibliographystyle{IEEEbib}
{\footnotesize
\bibliography{refs}}

\end{document}